\documentclass{bmvc2k}

\usepackage{tikz}
\usepackage{amsfonts}

\usepackage{booktabs}
\usepackage{wrapfig}
\usepackage{url}

\usepackage[colorlinks]{hyperref}

\title{SMPLitex: A Generative Model and Dataset for 3D Human Texture Estimation from Single Image}

\definecolor{amethyst}{rgb}{0.6, 0.4, 0.8}
\definecolor{darkpastelgreen}{rgb}{0.01, 0.75, 0.24}
\definecolor{amber}{rgb}{1.0, 0.75, 0.0}
\definecolor{cadmiumorange}{rgb}{0.93, 0.53, 0.18}
\definecolor{lawngreen}{rgb}{0.49, 0.99, 0.0}
\definecolor{limegreen}{rgb}{0.2, 0.8, 0.2}
\definecolor{neongreen}{rgb}{0.22, 0.88, 0.08}
\definecolor{amethyst}{rgb}{0.6, 0.4, 0.8}
\definecolor{darkpastelgreen}{rgb}{0.01, 0.75, 0.24}
\definecolor{greenbest}{RGB}{88,137,15}
\definecolor{redworst}{RGB}{137,15,27}

\newcommand{\REMOVE}[1]{{}}

\addauthor{Dan Casas}{dan.casas@urjc.es}{1}
\addauthor{Marc Comino-Trinidad}{marc.comino@urjc.es}{1}

\addinstitution{
 Universidad Rey Juan Carlos\\
 Madrid, Spain
}

\runninghead{Dan Casas and Marc Comino-Trinidad}{SMPLitex}

\begin{document}

\maketitle

\begin{abstract}
We propose SMPLitex, a method for estimating and manipulating the complete 3D appearance of humans captured from a single image.
SMPLitex builds upon the recently proposed generative models for 2D images, and extends their use to the 3D domain through pixel-to-surface correspondences computed on the input image.  
To this end, we first train a generative model for complete 3D human appearance, and then fit it into the input image by conditioning the generative model to the visible parts of subject.
Furthermore, we propose a new dataset of high-quality human textures built by sampling SMPLitex conditioned on subject descriptions and images. 
We quantitatively and qualitatively evaluate our method in 3 publicly available datasets, demonstrating that SMPLitex significantly outperforms existing methods for human texture estimation while allowing for a wider variety of tasks such as editing, synthesis, and manipulation.
\end{abstract}
\begin{center}
\href{https://dancasas.github.io/projects/SMPLitex}{dancasas.github.io/projects/SMPLitex} 
\end{center}

\vspace{-0.5cm}
\section{Introduction}
\vspace{-0.3cm}
\begin{figure}[h!]
    \centering
    \includegraphics[width=\textwidth]{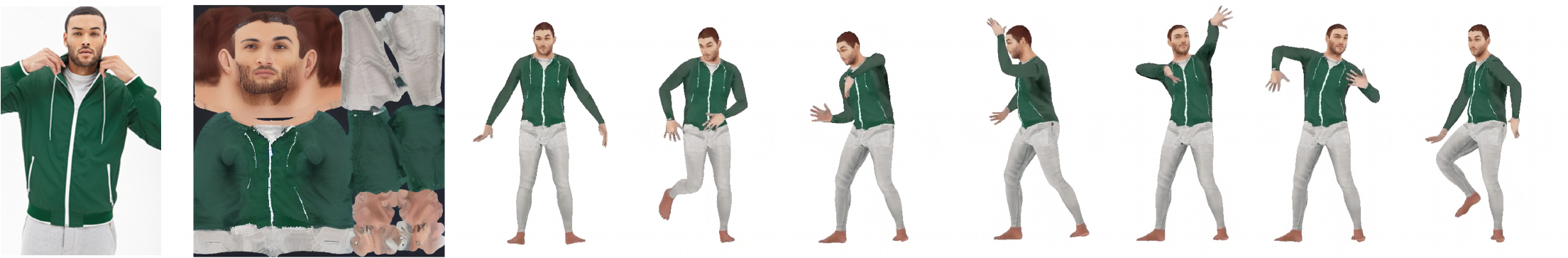}
    \caption{From a single image where a human is partly visible (left), SMPLitex automatically estimates a complete 3D texture map that can be applied to SMPL \cite{loper2015smpl} body mesh sequences.}
    \label{fig:label}
\end{figure}

Creating photorealistic 3D virtual humans is a long-standing goal in Computer Graphics and Computer Vision, with important applications in many areas including telecommunications, entertainment, online shopping, and medicine.
Among the many tasks required to produce digital humans, \textit{appearance synthesis} is the fundamental step to achieving photorealism.
To model virtual humans at scale, methods to create realistic-looking 3D human textures are needed.

Ideally, a method to model human appearance should be generative (to create new complete textures), easy to fit into partial observations (to recover textures from images), and compatible with traditional animation pipelines (to use the textures in popular commercial tools). 
Unfortunately, despite the impressive advances for digital humans, existing models for human synthesis typically focus on solving one of these challenges in isolation.
For example, recent methods have shown impressive results in \textit{posed image synthesis} tasks \cite{pumarola2018unsupervised,men2020controllable,sarkar2021style,sarkar2021humangan,grigorev2021stylepeople,lewis2021tryongan,fu2022stylegan} using generative strategies such as GANs~\cite{goodfellow2020generative,Karras2019stylegan2} or VAEs,  guided with 2D pose representations or text \cite{jiang2022text2human}.
However, these methods directly output camera-space 2D images of humans but do not generate a complete texture map.
This precludes their use in standard 3D animation pipelines, where UV texture maps for texturing 3D meshes are used.
Additionally, they are usually not ready to be used to recover textures from in-the-wild images.
Similarly, other works leverage neural rendering pipelines \cite{prokudin2021smplpix,noguchi2021neural,peng2021animatable,jiang2022neuman,weng2022humannerf} to generate view-dependent posed avatars but do not output 3D texture maps either. 

Alternatively, and closer to ours, other works focus on \textit{complete texture estimation} from single image~\cite{lazova2019360,kanazawa2018learning,neverova2018dense,wang2019reidsupervised,xu2021texformer,zhao2020human} and are able to recover 3D texture maps from casual images.
Most of these methods used CNN architectures \cite{lazova2019360,wang2019reidsupervised} to infer the complete 3D texture map from a single image, sometimes using multi-view supervision \cite{zhao2020human} or transformer-based architectures \cite{xu2021texformer}. 
However, they are limited to generating textures from images, which precludes their use in applications that require the synthesis of unseen virtual humans (\textit{e.g.}, via text prompts).
Most importantly, they typically output low-detail textures due to the limited expressivity of the latent space of the network.
Additionally, we argue that these undesired properties prevent the use of existing methods to build large and high-quality public datasets of 3D human textures which, we believe, is a major shortfall in the field.

In this paper, we address these limitations and propose SMPLitex, a generative method for complete 3D human texture synthesis.
SMPLitex enables the estimation of 3D human textures from single images that can be directly applied to SMPL meshes as shown in Figure \ref{fig:label}.
Additionally, since it is a generative method, it also allows for the synthesis of new textures via text prompts or image editing, as shown in Figures \ref{fig:my_label} and \ref{fig:smplitex-dataset}.
Under the hood, SMPLitex leverages the recently proposed diffusion models for 2D image synthesis \cite{ramesh2022hierarchical,rombach2022high} built from a hierarchy of denoising processes.
Despite the impressive results of such models in 2D  tasks \cite{lugmayr2022repaint,chung2022improving,ho2022video,ho2022imagen,voleti2022MCVD}, 
extending diffusion models to 3D humans requires addressing two important challenges:
spatial regularization, to enforce multi-view consistency; and 3D awareness, to enable 2D-to-3D tasks.
Our key observation is that, for human-related tasks, a proxy of the 3D geometry visible in any image is coarsely modeled with human body models such as SMPL \cite{loper2015smpl}, which opens the door to 2D diffusion models for 3D humans.
To this end, we first learn a domain-specific diffusion model that is trained to generate \textit{unwrapped} 3D textures of humans, which implicitly learns multi-view consistency. 
Then, to provide the 3D awareness required for 2D-to-3D tasks such as 3D texture from monocular input, we estimate pixel-to-surface correspondences \cite{guler2018densepose} to project image pixels to an incomplete 3D texture map. 
Leveraging the 2D structure implicitly enforced in our domain-specific diffusion model, we are able to inpaint the incomplete 3D texture map.

We demonstrate that the proposed model, SMPLitex, outperforms state-of-the-art methods \cite{lazova2019360,xu2021texformer} for 3D human texture estimation in 3 publicly available datasets  \cite{zheng2015scalable,jiang2022text2human,liuLQWTcvpr16DeepFashion}, generating high-resolution texture maps.
Additionally, we exploit the generative capabilities of the proposed model to create a new dataset of high-quality 3D textures.
Our new dataset overcomes the quality, diversity, and number of samples of existing datasets such as SURREAL \cite{varol17surreal}.
This paves the way for new data-driven models that require photorealistic human data such as 3D pose estimation, 3D human reconstruction, and neural rendering.

In summary, our main contributions are:
\begin{itemize}
\item SMPLitex, a new generative model for 3D human textures that can be used as a drop-in replacement for textures in any SMPL-based pipeline.
\item A novel diffusion-based method to infer 3D human textures from single RGB input.
\item A new dataset of high-quality 3D human textures that significantly surpasses the detail, diversity, and size of existing datasets.  
\end{itemize}

\section{Related Work}
\textbf{Texture recovery from single image.}
This group of methods attempts to recover the complete appearance of humans \cite{lazova2019360,neverova2018dense,wang2019reidsupervised,xu2021texformer,zhao2020human,cha2023generating,alldieck2022},
faces \cite{saito2017photorealistic} or 
category-specific objects \cite{kanazawa2018learning,mir2020learning} from a single image.
To tackle the challenges arising from such ill-posed problem, a variety of learning-based solutions relying on neural networks have been proposed, including the use of multi-view supervision~\cite{zhao2020human,neverova2018dense}, transformer-based architectures \cite{xu2021texformer}, and differentiable rendering~\cite{kanazawa2018learning,wang2019reidsupervised}.

Similar to ours, some works frame the texture recovery problem as image inpainting problem \cite{lazova2019360,neverova2018dense}.
For example, Lazova \textit{et al.}~\cite{lazova2019360} computes pixel-to-surface correspondences from the input image and project the pixel information into a partial texture map. The incomplete texture is then inpainted using a GAN-based network, generating a complete 3D texture of the input image.
Neverova \textit{et al.} \cite{neverova2018dense} also train a partial texture inpainting network to infer the occluded parts of the body, which is supervised by multi-view ground truth data.
Instead of inpainting a texture map, other methods directly predict the full texture given the input image~\cite{xu2021texformer,zhao2020human,wang2019reidsupervised,xu20213tpami}.
For example, Xu \textit{et al.}~\cite{xu2021texformer} map 3D coordinates of a human body mesh to a UV texture map which, in combination with a 2D part segmentation image, is converted to a texture map using a transformer-based network.
Wang \textit{et al.}~\cite{wang2019reidsupervised} utilize a distance metric based on a re-identification loss to learn to generate texture maps given a dataset of images of humans taken from different viewpoints and their corresponding 3D pose.
Similarly, Zhao \textit{et al.}~\cite{zhao2020human} add part-based segmentation and enforce cross-view consistency to learn to generate complete texture maps.   
Despite the impressive results, these methods typically output low-detail textures due to the limited expressivity of the latent space of the networks used.
Furthermore, they are also limited to generating textures from images, which precludes their use in applications that require the synthesis of new textures or their manipulation, for example, via text prompts.

Closely related to the texture estimation task are the method that aim to reconstruct 3D humans and appearance from single image \cite{he2021arch,alldieck2022,zheng2020pamir,li2022avatarcap,natsume2019siclope}.
These methods produce high-fidelity 3D reconstructions, including fine geometric details, but the estimated appearance is usually not explicitly baked into a consistent UV texturemap. Furthermore, they do not enable the synthesis of unseen appearances.

\vspace{8pt}
\noindent\textbf{Posed image synthesis.}
Instead of predicting the complete 3D texture map, this group of methods aim at generating images of posed humans \cite{pumarola2018unsupervised,grigorev2019coordinate,men2020controllable,sarkar2021style,sarkar2021humangan,grigorev2021stylepeople,lewis2021tryongan,fu2022stylegan,jiang2022text2human}, hence only requiring to synthesize the visible parts of the subject.  
Under the hood, these methods use generative strategies, such as GANs~\cite{goodfellow2020generative,Karras2019stylegan2} or VAEs, guided with 2D pose representations \cite{pumarola2018unsupervised}, dense surface correspondences \cite{grigorev2019coordinate,albahar2021pose}, or text prompts \cite{jiang2022text2human} to describe the target pose.
For example, Sarkar \textit{et al.}~\cite{sarkar2021style} encode the partial (\textit{i.e.}, visible) UV-space appearance to a global latent vector to modulate a StyleGAN2~\cite{Karras2019stylegan2} image generator. Similarly, AlBahar \textit{et al.}~\cite{albahar2021pose} inpaint a correspondence field and transfer local surface details to the target pose. 

Despite the high-quality results of these methods, most of them do not output an explicit texture map that can be used in 3D animation pipelines. This limits their use to purely 2D image synthesis use cases. \cite{liu2021tvcg,habermann2021} are notable exceptions and predict explicit dynamic texture maps of the subject for pose synthesis, but require per-subject retraining.

\vspace{8pt}
\noindent\textbf{Avatars from text.}
Also related to ours are the methods that aim at synthesizing humans from text descriptions~\cite{jiang2022text2human,hong2022avatarclip,cheong2022kpe,zhang2023avatarverse,kolotouros2023dreamhuman}.
These methods combine recent image generative models \cite{fu2022stylegan} based on GANs, VAEs, or diffusion models, with large vision-language pre-trained models such as CLIP~\cite{radford2021learning} to condition the output.
For example, AvatarCLIP~\cite{hong2022avatarclip} is able to generate and animate 3D textured humans directly from text,
and Text2Human~\cite{jiang2022text2human} synthesizes high-quality 2D posed humans given detailed outfit descriptions.
Other works focus on more general objects \cite{jain2021dreamfields,khalid2022clipmesh,michel2022text2mesh}.
However, they do not generate consistent texture maps, and cannot easily be fitted into partial observations to estimate textures from images.

\section{3D Human Texture Estimation}
 Given an input RGB image $\mathbf{x}$, where a person is visible or partially visible, our goal is to estimate a 3D texture map $\mathbf{u}$ that encodes the complete appearance of the subject.
This is a challenging ill-posed problem because (1) the 3D geometry of the scene (\textit{i.e.}, the 3D surface of the person) is unknown; and (2) many parts of the image suffer from natural occlusions and self-occlusions (\textit{i.e.}, not all surface points are visible).

To address this problem, a common approach \cite{lazova2019360,xu2021texformer} is, first, to use a coarse geometry proxy \cite{loper2015smpl} and infer pixel-to-surface correspondences \cite{guler2018densepose} to build an incomplete texture map; and later use an image-to-image translation framework to inpaint or estimate the incomplete texture.
However, the limited expressivity of existing methods leads to low-resolution inpainted textures that lack detail.
Furthermore, the resulting models cannot be manipulated with text prompts, which limits their applicability to image-to-image tasks.

\begin{figure*}
    \centering
    \includegraphics[width=\textwidth]{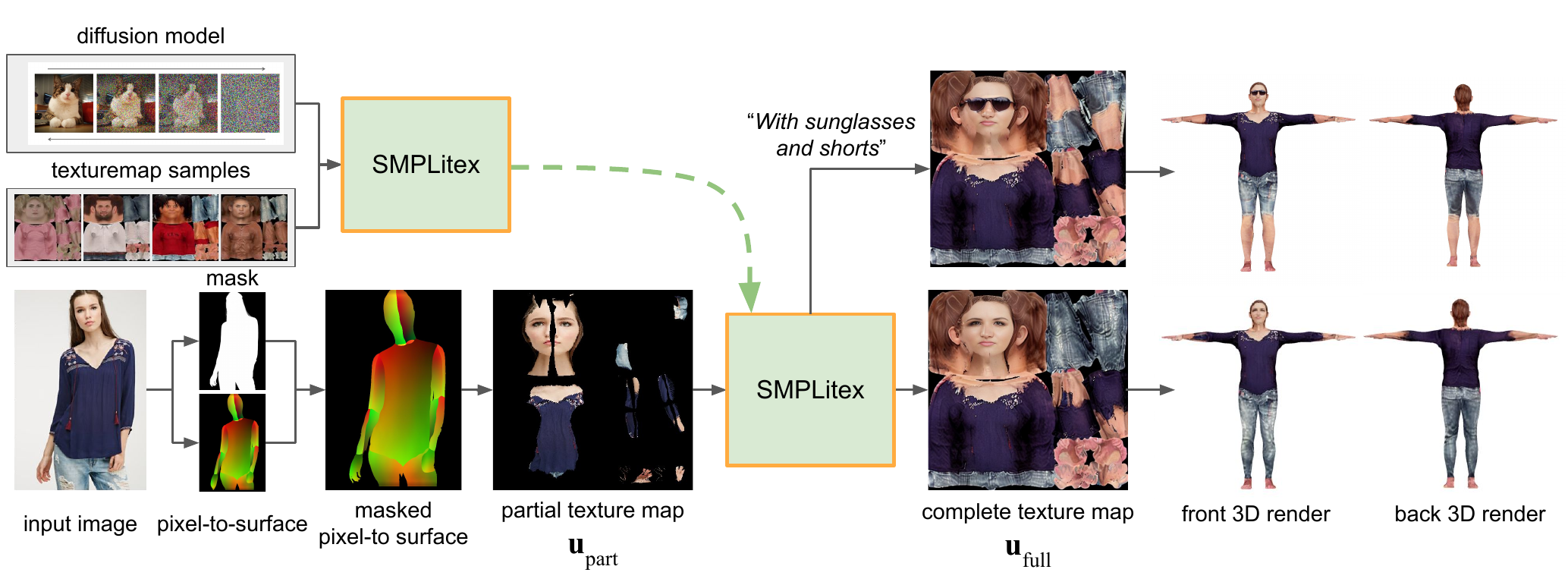}
    \caption{Overview of the proposed approach for texture estimation from a single image.}
    \label{fig:my_label}
\end{figure*}

In contrast, we propose a pipeline based on image diffusion models \cite{rombach2022high} that is capable of recovering high-quality textures, and additionally, it naturally allows for text-based manipulations. 
Figure \ref{fig:my_label} presents a visualization of our pipeline. 
In the rest of this section, we first describe how we formulate and train our generative model (Section \ref{sec:generative-model}), and then describe how we leverage it to estimate complete 3D textures from in-the-wild monocular RGB images (Section \ref{sec:texture-estimation}).
\subsection{Generative 3D Human Textures}
\label{sec:generative-model}
\subsubsection{Background}
SMPLitex uses a diffusion model as a generative backbone.
Diffusion models are a type of generative model that gradually remove noise from an image to learn the distribution space $p_{\theta}$~\cite{ho2020denoising}. 
To accomplish this, the model starts with a sampled Gaussian noise and performs a step-by-step denoising process over $T$ time steps until it produces a final, noise-free image.
During each step of the denoising process, the diffusion model generates noise $\epsilon_t$ that is used to create an intermediate denoised image $\mathbf{x}_{t}$. The initial noise $\mathbf{x}_T$ corresponds to the final image, while the fully denoised image $\mathbf{x}_0$ corresponds to the starting image. 
This denoising process is typically modeled as a Markov transition probability as follows
\begin{align}
    p_{\theta}(\mathbf{x}_{T:0}) = p(\mathbf{x}_T)\prod_{t=T}^{1}p_{\theta}(\mathbf{x}_{t-1} | \mathbf{x}_t)
\end{align}

To improve the efficiency of such diffusion models, \textit{latent} diffusion models (LDM)~\cite{rombach2022high} operate on a lower dimensional latent space $\mathbf{z}$ pre-trained using a variational autoencoder. 
During training, for an image $\mathbf{x}$, noise is added to the encoded image $\mathbf{z} = \mathcal{E}(\mathbf{x})$, where $\mathcal{E}$ is the pre-trained encoder, leading to $\mathbf{z}_t$ where the noise level increases with $t$. 
Analogous to the original diffusion model, the LDM process can be seen as a sequence of denoising models with shared parameters $\theta$ that learn to predict a noisy image $\epsilon_\theta(\mathbf{z}_t, c, t)$, where $t$ is the timestep and $c$ a text condition. LDMs are trained by minimizing the loss term
\begin{equation}
\mathbb{E}_{\mathbf{z}_t\in\mathcal{E}(x) \text{, } t \text{, } c \text{, } \epsilon\sim\mathcal{N}(0,1)} [| \epsilon - \epsilon_\theta(\mathbf{z}_t, c, t)|_2^2]
\label{eq:ldm}
\end{equation}

Once trained, the LDM model can generate new samples following the diffusion process in reverse mode, iteratively predicting the noise to be removed from a randomly sampled Gaussian noise (potentially conditioned on text and time step).   

\subsubsection{Diffusion Model for 3D Humans}
Many extensions of the original LDM \cite{rombach2022high} model have been recently proposed for a wide variety of tasks and data modalities, including image inpainting~\cite{lugmayr2022repaint,chung2022improving,suvorov2022resolution} and video  generation~\cite{ho2022video,ho2022imagen,voleti2022MCVD}.
In this work, we look into how LDM can be adapted to the specific case of 3D human appearance.
However, naively extending LDMs to 3D domains is  challenging since multi-view consistency cannot be guaranteed. 
A few recent works demonstrate promising results in text-to-3D~\cite{poole2023dreamfusion} or novel view synthesis~\cite{watson2023novel,bautista2022gaudi} tasks,
but articulated objects and high-resolution images remain a challenge.

Our key observation is that, for the specific case of 3D humans, these limitations can be circumvented by using a 3D-to-2D parametrization of the surface (\textit{i.e.}, a UV map of the mesh surface).
By working directly on the UV map, we are able to encode the 3D appearance of the human directly on a 2D image, opening the door to the potential of LDMs models for appearance synthesis, inpainting, and manipulation.

Therefore, we propose to leverage the highly expressive LDM  proposed by Rombach \textit{et al.}~\cite{rombach2022high}, and fine-tune it to encode UV textures of humans.
The main underlying challenge of this task is how to ensure that the expressivity of the original model is preserved (\textit{i.e.}, the fine-tuned model is able to generate detailed and rich images that are not in the fine-tuning training set) while satisfying the spatial constraints inherent in the UV textures (\textit{i.e.}, the 2D-to-3D parameterization).
Crucially, recent text-to-image models~\cite{ruiz2023dreambooth,hu2021lora} address an analogous problem for model personalization, where the goal is to learn to generate images with a specific style or for a particular subject.
Under the hood, these models use a class-specific prior preservation loss that enables the synthesis of the target subject or style in arbitrary scenarios, just using an extremely reduced set of training samples.
Inspired by this, SMPLitex leverages the work of Ruiz \textit{et al.} \cite{ruiz2023dreambooth} to fine-tune the model of Rombach \textit{et al.}~\cite{rombach2022high} such that it is constrained to synthesize SMPL UV texture maps.
In practice, we use 10 UV texture maps for SMPL from \cite{alldieck2018video,lazova2019360} to fine-tune the model \cite{rombach2022high} available at \cite{hugging} for 1,500 iterations. At inference time, our results use 50 denoising steps and a classifier free guidance (CGF) of 2.0.

\subsection{Human Texture Estimation from Single Image}
\label{sec:texture-estimation}
The diffusion model described in Section \ref{sec:generative-model} enables the synthesis of high-quality UV texture maps of humans. 
As discussed above, the diffusion model is sampled conditioned to the time step, such that it removes the noise at time $t$, but it also allows for additional conditioning signals (\textit{e.g.}, condition text $c$ in Equation \ref{eq:ldm}). 
Our key intuition is that to enable the estimation of 3D human appearance from a single image, we can condition the synthesis of an LDM model for human appearance \textit{to the visible parts of the subject} in the input image.
In other words, we are interested in fitting SMPLitex into natural images.

To this end, similar to \cite{lazova2019360}, we leverage the fact that we are under the assumption of modeling 3D humans and use state-of-the-art pixel-to-surface correspondence models to compute a partial texture map $\mathbf{u}_{\text{part}}$.
More specifically, given an input image $\mathbf{x}$, we estimate pixel-to-surface correspondences $\mathbf{d}$ \cite{neverova2018dense} and project the pixels of $\mathbf{x}$ with assigned surface correspondences to a partial UV map $\mathbf{u}_{\text{part}}$, which will be used as a conditional signal.

However, a naive use of the pixel-to-surface correspondence $\mathbf{d}$ can potentially lead to undesired partial UV maps since $\mathbf{d}$ typically coarsely estimates the foreground silhouette.
This can lead to background pixels projected into the UV map, which significantly degenerates the condition image.  
We mitigate this issue by computing an accurate human silhouette $\mathbf{s}$~\cite{chen2022sghm}, that we use to mask the pixel-to-surface image $\mathbf{d}$.
More formally, we compute our partial UV map as
\begin{equation}
    \mathbf{u}_{\text{part}} =  \Pi(\mathbf{x},\mathbf{d} \odot \mathbf{s})
\end{equation}
where $\odot$ is the Hadamard product, $\Pi$ is the operator that projects all pixels $p \in \mathbf{x}$ to their corresponding surface coordinate according to the map $\mathbf{d} \odot \mathbf{s}$. 

With the partial texture map $\mathbf{u}_{\text{part}}$ computed, we can infer the complete texture map $\mathbf{u}_{\text{full}}$ of the input image $\mathbf{x}$ by just sampling the diffused model described in Section \ref{sec:generative-model} using $\mathbf{u}_{\text{part}}$ as an additional conditioning signal.

\section{SMPLitex Dataset}
\begin{figure*}
\includegraphics[trim={50 100 590 100},clip,width=0.1\textwidth]{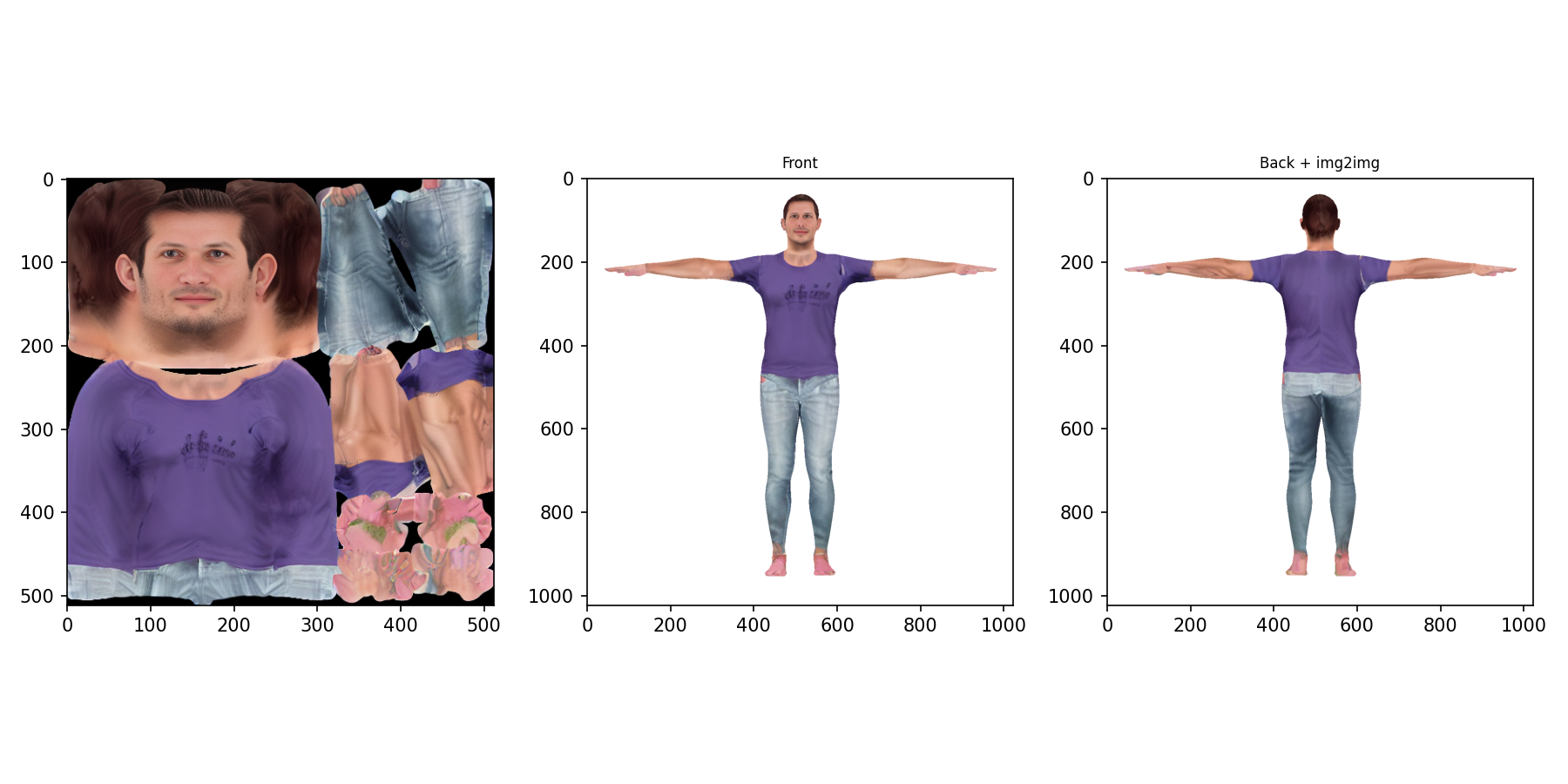}
\includegraphics[trim={330 100 310 100},clip,width=0.1\textwidth]{images/dataset/a_sks_texturemap_of_a_Magician-batch-002_sample-003-debug.png}
\includegraphics[trim={611 100 30 100},clip,width=0.1\textwidth]{images/dataset/a_sks_texturemap_of_a_Magician-batch-002_sample-003-debug.png}
\hspace{0.3cm}
\includegraphics[trim={50 100 590 100},clip,width=0.1\textwidth]{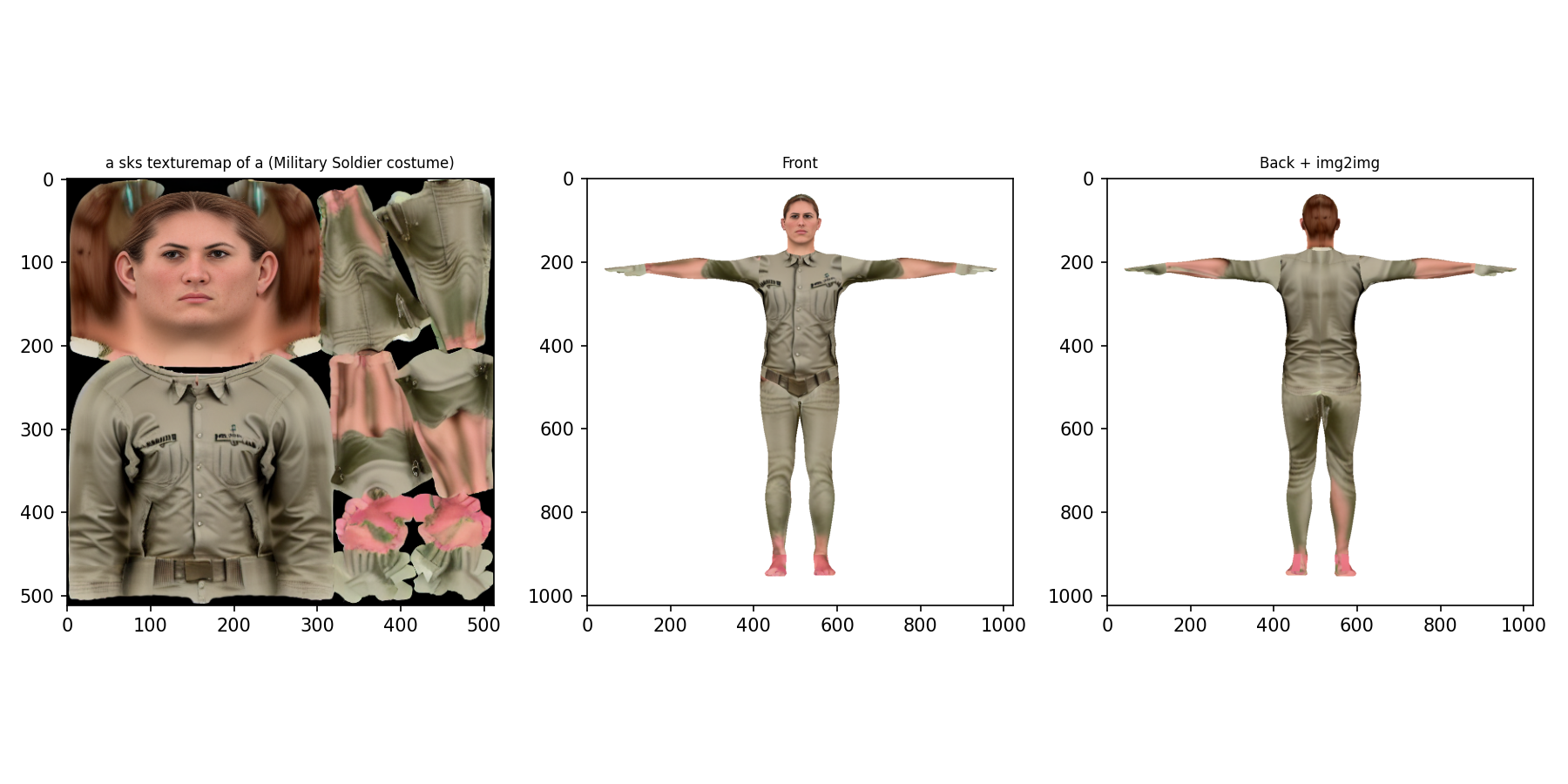}
\includegraphics[trim={330 100 310 100},clip,width=0.10\textwidth]{images/dataset/Military-Soldier-costume_cfg-2.5_batch-003_sample-000_pants-002-debug.png}
\includegraphics[trim={611 100 30 100},clip,width=0.1\textwidth]{images/dataset/Military-Soldier-costume_cfg-2.5_batch-003_sample-000_pants-002-debug.png}
\hspace{0.3cm}
\includegraphics[trim={50 100 590 100},clip,width=0.1\textwidth]{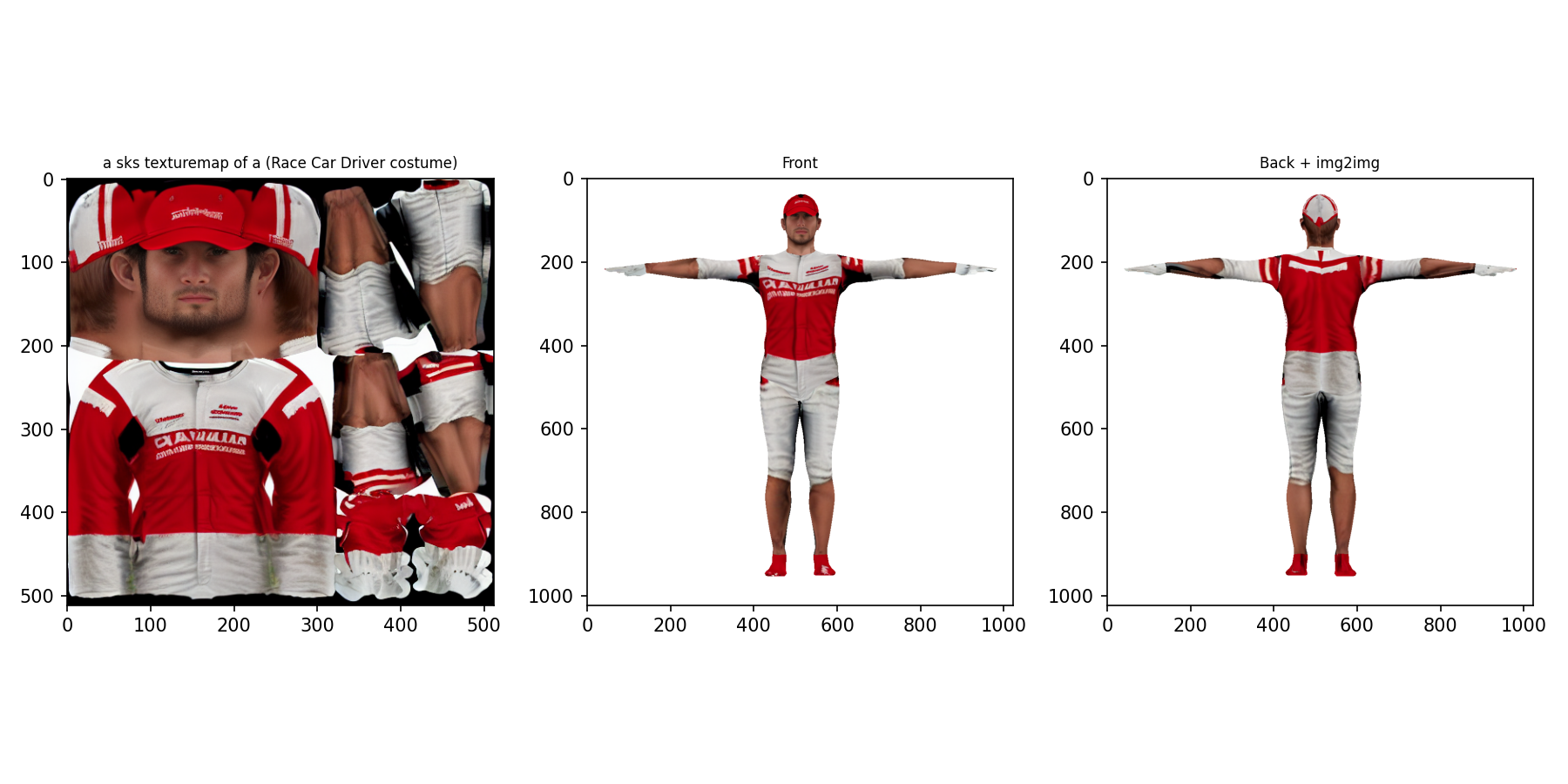}
\includegraphics[trim={330 100 310 100},clip,width=0.1\textwidth]{images/dataset/Race-Car-Driver-costume-batch-000_sample-001-debug.png}
\includegraphics[trim={611 100 30 100},clip,width=0.1\textwidth]{images/dataset/Race-Car-Driver-costume-batch-000_sample-001-debug.png}
\\[-0.48cm]

\hspace{1.1cm}
\footnotesize{\textit{"Casual outfit"}}
\hspace{2.1cm}
\footnotesize{\textit{"Military soldier costume"}}
\hspace{1.6cm}
\footnotesize{\textit{"Race car driver"}}
\\[0.2cm]
\includegraphics[trim={50 100 590 100},clip,width=0.1\textwidth]{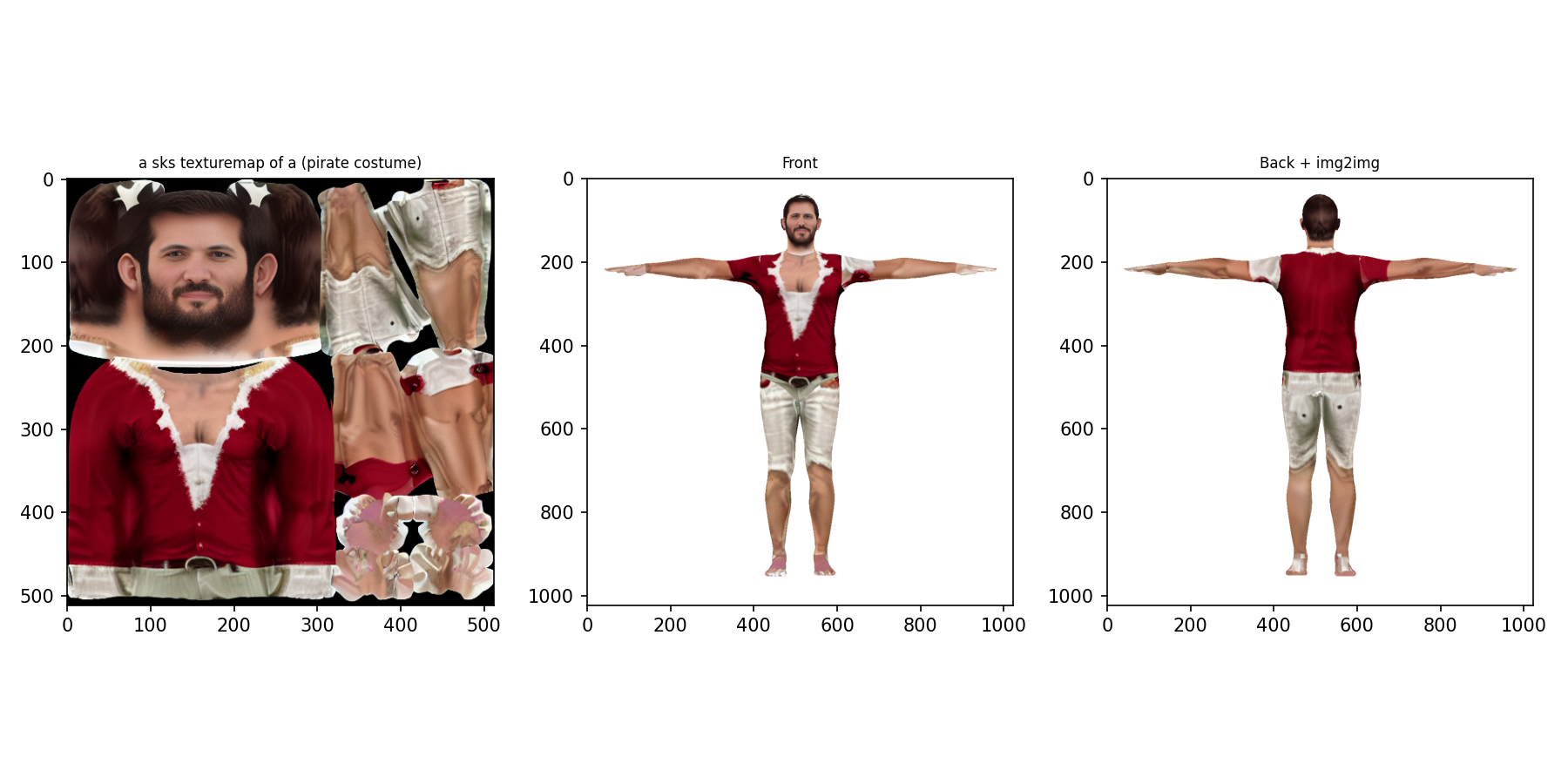}
\includegraphics[trim={330 100 310 100},clip,width=0.1\textwidth]{images/dataset/pirate-costume-batch-006_sample-003-debug.png}
\includegraphics[trim={611 100 30 100},clip,width=0.1\textwidth]{images/dataset/pirate-costume-batch-006_sample-003-debug.png}
\hspace{0.3cm}
\includegraphics[trim={50 100 590 100},clip,width=0.1\textwidth]{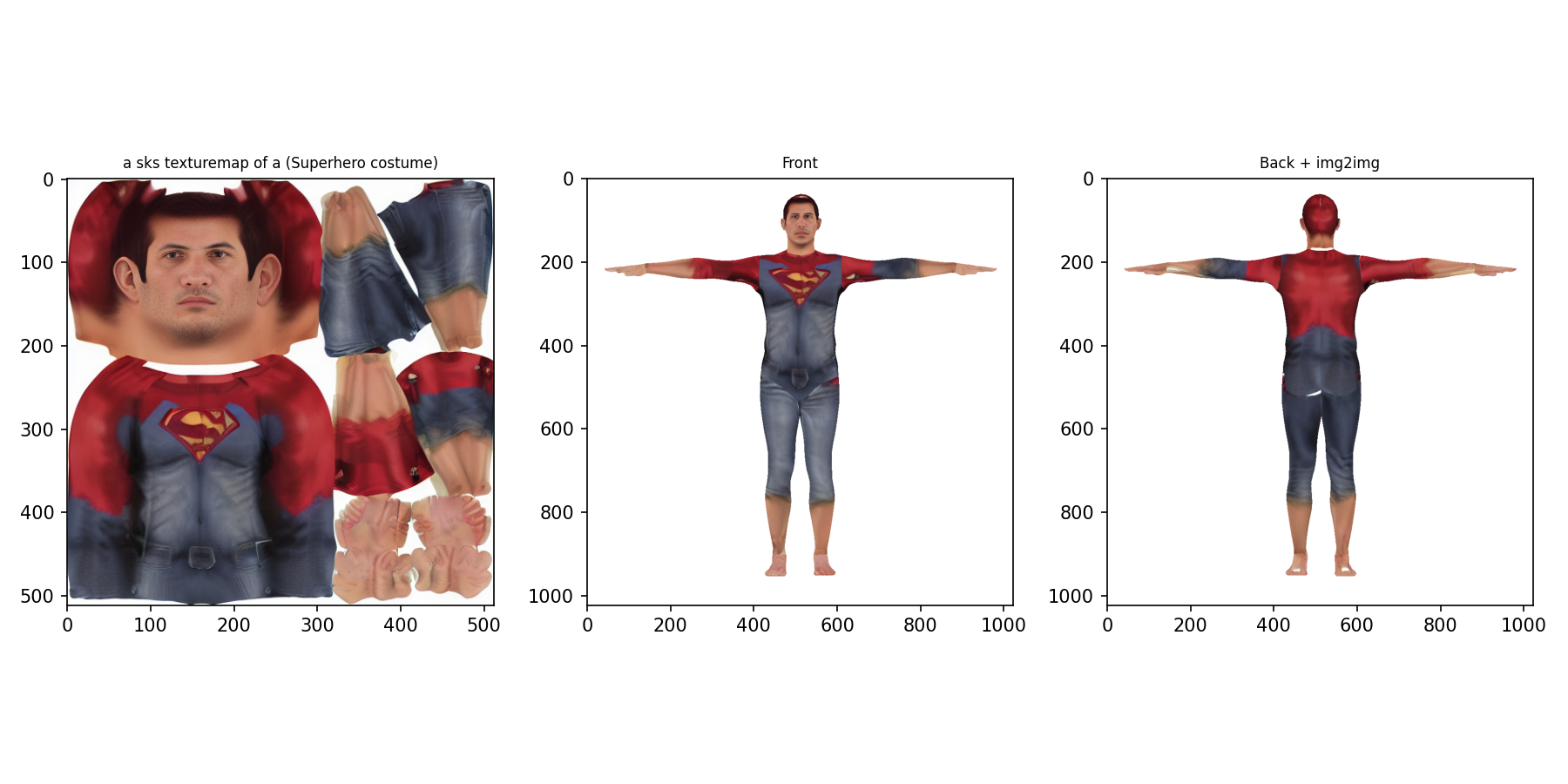}
\includegraphics[trim={330 100 310 100},clip,width=0.10\textwidth]{images/dataset/Superhero-costume_cfg-2.0_batch-001_sample-002-debug.png}
\includegraphics[trim={611 100 30 100},clip,width=0.1\textwidth]{images/dataset/Superhero-costume_cfg-2.0_batch-001_sample-002-debug.png}
\hspace{0.3cm}
\includegraphics[trim={50 100 590 100},clip,width=0.1\textwidth]{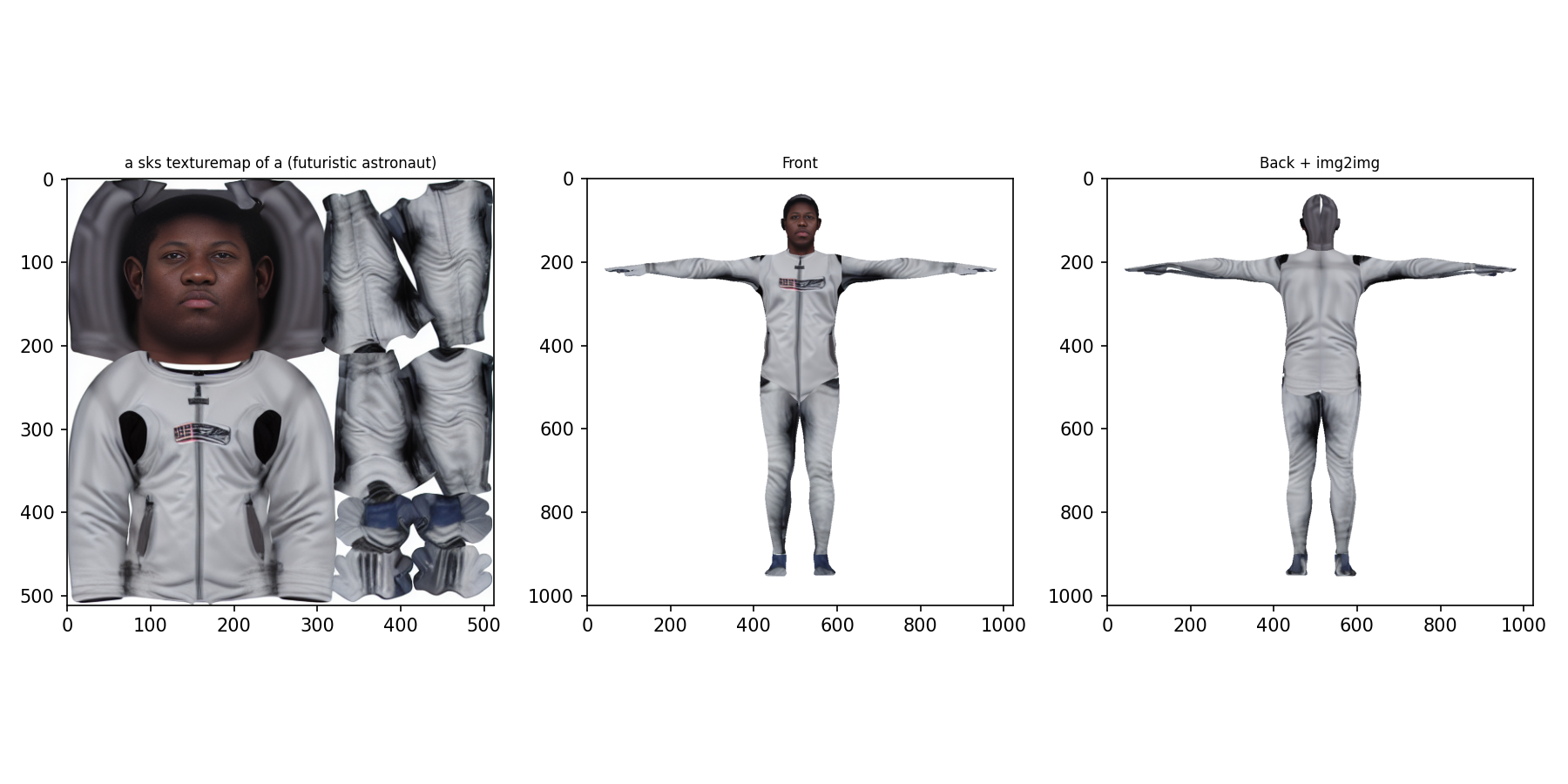}
\includegraphics[trim={330 100 310 100},clip,width=0.1\textwidth]{images/dataset/futuristic-astronaut-batch-004_sample-001-debug.png}
\includegraphics[trim={611 100 30 100},clip,width=0.1\textwidth]{images/dataset/futuristic-astronaut-batch-004_sample-001-debug.png}
\\[-0.4cm]

\hspace{1.0cm}
\footnotesize{\textit{"Pirate costume"}}
\hspace{2.3cm}
\footnotesize{\textit{"Superhero costume"}}
\hspace{1.8cm}
\footnotesize{\textit{"Futuristic astronaut"}}
\\[0.2cm]
\includegraphics[trim={50 100 590 100},clip,width=0.1\textwidth]{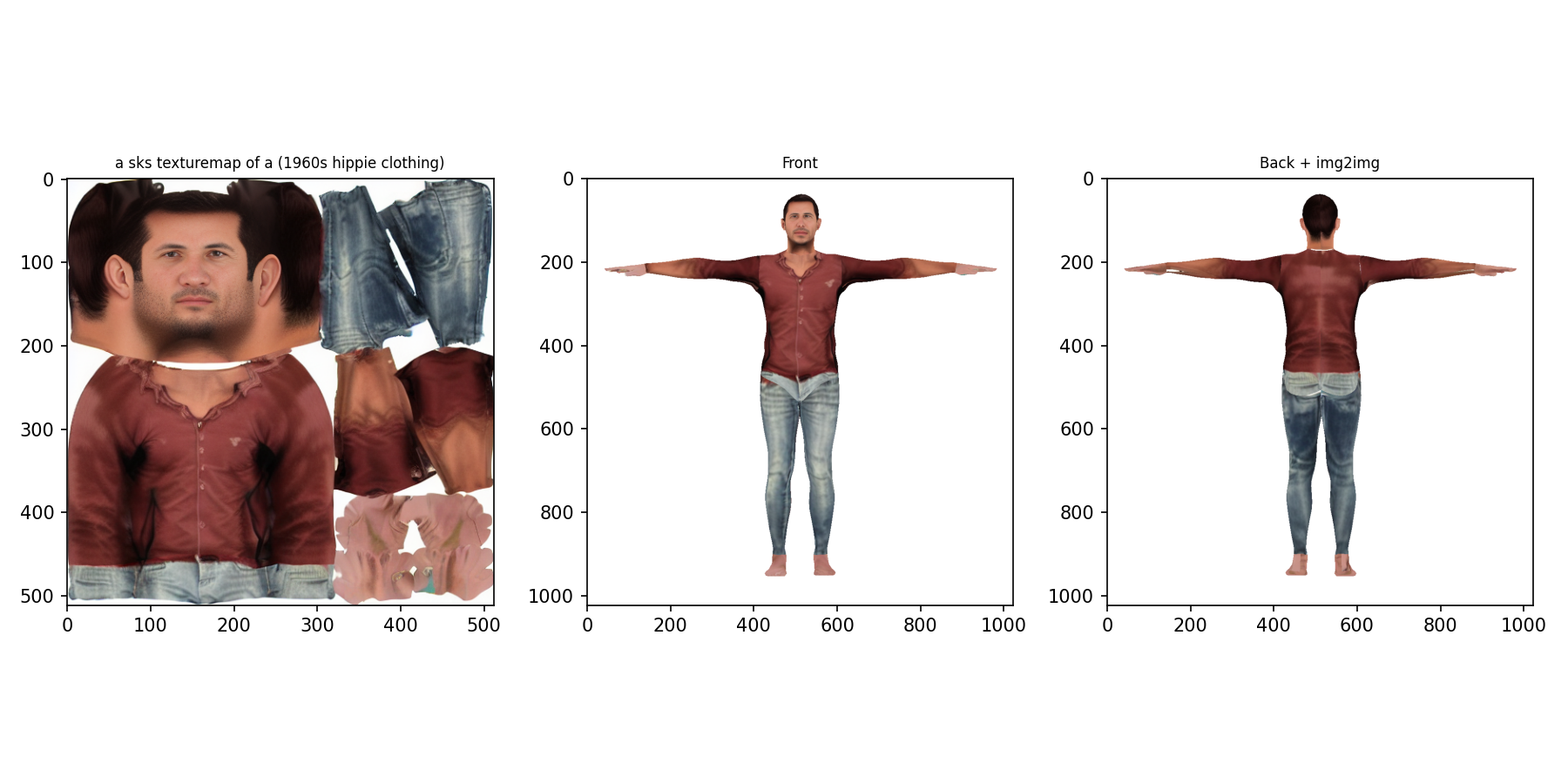}
\includegraphics[trim={330 100 310 100},clip,width=0.1\textwidth]{images/dataset/1960s-hippie-clothing_cfg-2.0_batch-001_sample-001_pants-002-debug.png}
\includegraphics[trim={611 100 30 100},clip,width=0.1\textwidth]{images/dataset/1960s-hippie-clothing_cfg-2.0_batch-001_sample-001_pants-002-debug.png}
\hspace{0.3cm}
\includegraphics[trim={50 100 590 100},clip,width=0.1\textwidth]{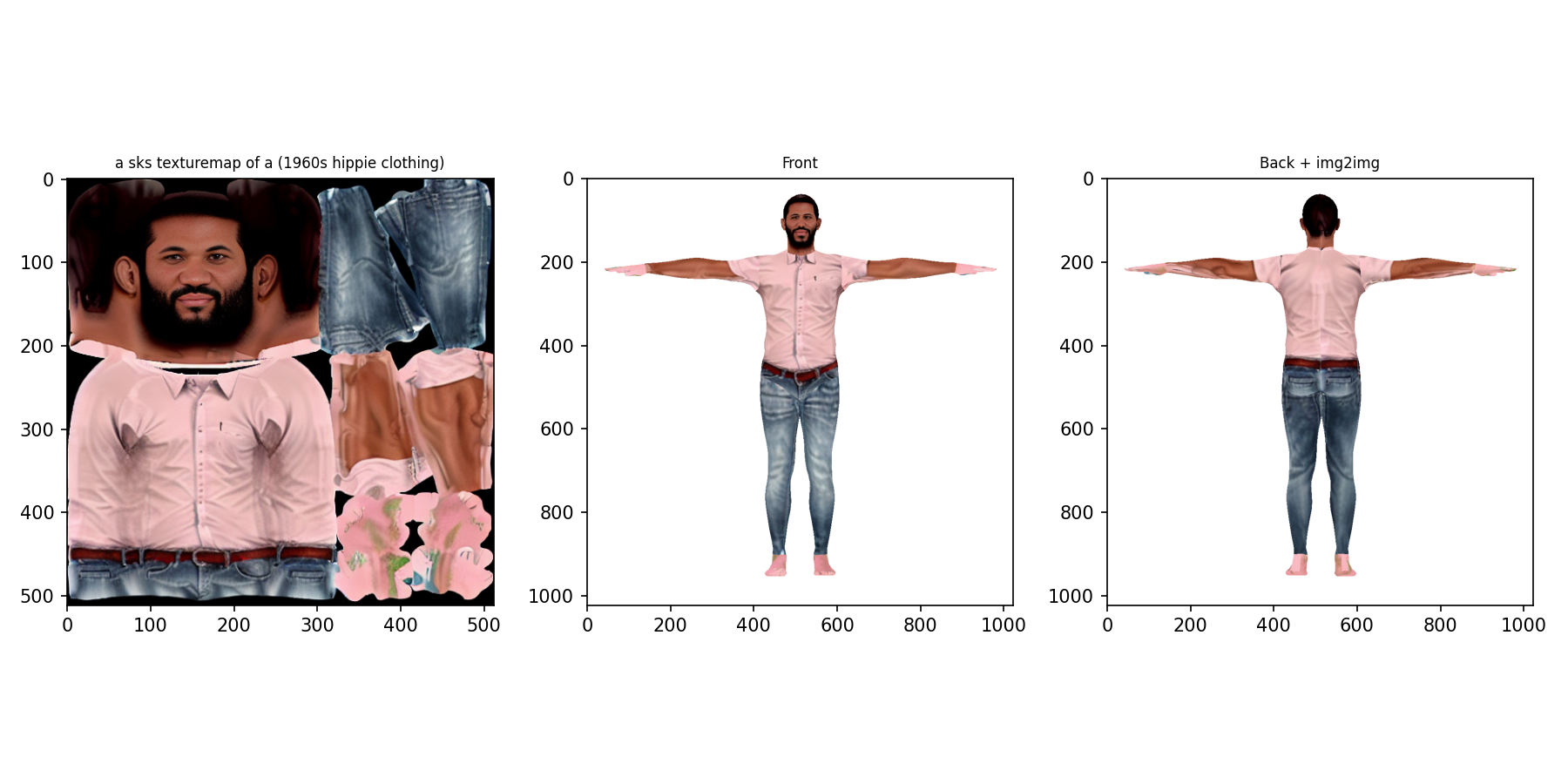}
\includegraphics[trim={330 100 310 100},clip,width=0.10\textwidth]{images/dataset/1960s-hippie-clothing_cfg-2.75_batch-004_sample-003_pants-004-debug.png}
\includegraphics[trim={611 100 30 100},clip,width=0.1\textwidth]{images/dataset/1960s-hippie-clothing_cfg-2.75_batch-004_sample-003_pants-004-debug.png}
\hspace{0.3cm}
\includegraphics[trim={50 100 590 100},clip,width=0.1\textwidth]{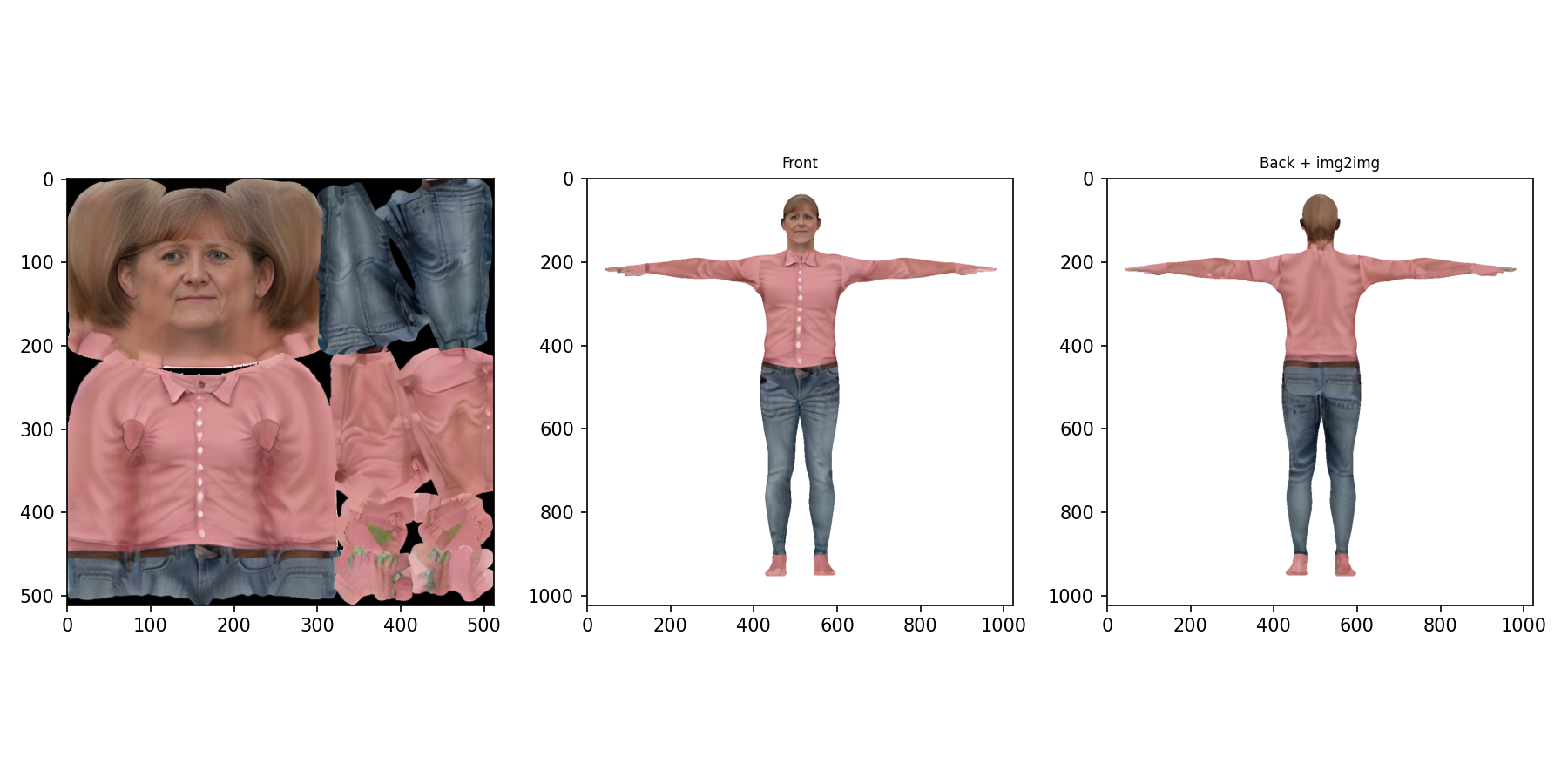}
\includegraphics[trim={330 100 310 100},clip,width=0.1\textwidth]{images/dataset/a_sks_texturemap_Angela_Merkel-batch-003_sample-003-debug.png}
\includegraphics[trim={611 100 30 100},clip,width=0.1\textwidth]{images/dataset/a_sks_texturemap_Angela_Merkel-batch-003_sample-003-debug.png}
\\[-0.4cm]

\hspace{1.1cm}
\footnotesize{\textit{"Hippie outfit"}}
\hspace{2.6cm}
\footnotesize{\textit{"Business outfit"}}
\hspace{2.1cm}
\footnotesize{\textit{"Angela Merkel, shirt"}}
\caption{SMPLitex texture samples, side-by-side to their corresponding front and back SMPL render. Notice that these textures are not used for training SMPLitex, instead, they are part of the dataset that we built by arbitrarily sampling our model. 
We believe such high-quality textures will open the door to new possibilities in the area of virtual humans.}
\label{fig:smplitex-dataset}
\end{figure*}

Taking advantage of the generative capabilities of the appearance model described in Section \ref{sec:generative-model}, we build a dataset of high-quality textures by simply sampling the latent space.
More specifically, we use text conditioning prompts describing arbitrary outfit combinations, costumes, sports apparel, job titles, and facial characteristics.
Figure \ref{fig:smplitex-dataset} depicts 9 samples of the dataset, showcasing a large variety of garment types, outfits, identities, and facial details.
In total, the SMPLitex dataset consists of 100 curated textures, see the supplementary material for more details.
To clarify, on top of the dataset we will also release the trained SMPLitex model to generate new arbitrary samples, and it is the core component of our method to recover textures from single images. However, we believe that releasing a fixed set of curated textures will benefit future research.

\section{Results and Evaluation}
We qualitatively and quantitatively evaluate our method on 3 publicly available datasets \cite{jiang2022text2human,xu2021texformer,tao2021function4d} and demonstrate that our results compare favorably with state-of-the-art methods for texture estimation.
In contrast to existing methods, SMPLitex can deal with both low and high-resolution imagery, and it is robust to multi-view consistency metrics.

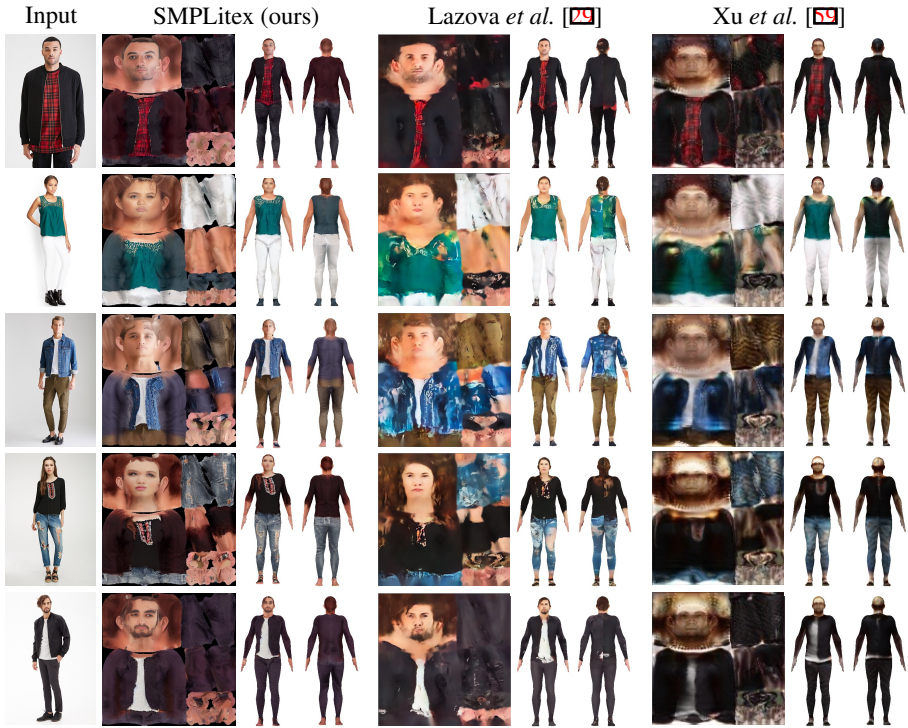
\begin{figure*}
\centering
\resizebox{0.95\columnwidth}{!}{%
\begin{tikzpicture}
    \node (img) at (0,0) {\includegraphics[width=\textwidth]{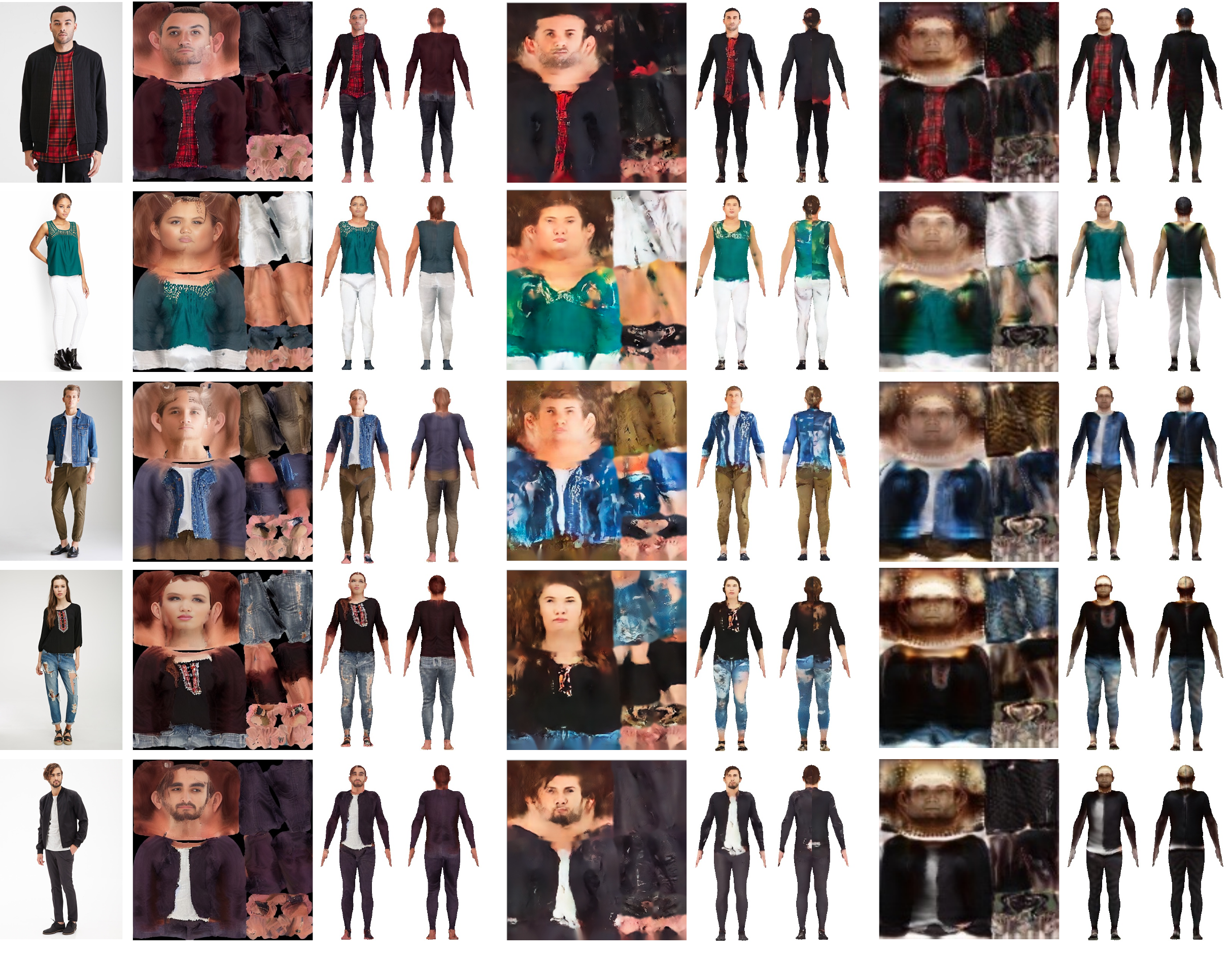}};
    \node (input)[above, align=left, xshift = -165pt, yshift = -5pt] at (img.north){Input};
    \node (ours) [above, align=left, xshift = -90pt, yshift = -5pt] at (img.north){SMPLitex (ours)};
    \node (lazova) [above, align=left, xshift = +23pt, yshift = -5pt] at (img.north){Lazova \textit{et al.}~\cite{lazova2019360}};
    \node (texformers) [above, align=left, xshift = +130pt, yshift = -5pt] at (img.north){Xu \textit{et al.}~\cite{xu2021texformer}};
\end{tikzpicture}
}
\caption{Qualitative comparison with state-of-the-art methods \cite{lazova2019360,xu2021texformer} in the DeepFashion-MultiModal dataset~\cite{jiang2022text2human}. SMPLitex clearly outperforms the texture quality of previous methods, recovering fine details such as garment wrinkles and facial attributes.}
\label{fig:comparison-with-lazova-texformer}
\end{figure*}

\vspace{-10pt}
\paragraph*{Evaluation on DeepFashion-MultiModal~\cite{jiang2022text2human}.}
This dataset consists of a large collection of high-resolution fashion images ($750\times1101$ pixels), where subjects wear a wide variety of clothing styles.
In Figure \ref{fig:comparison-with-lazova-texformer} we present our qualitative results on this dataset, and compare them to the results of the state-of-the-art closest methods \cite{lazova2019360,xu2021texformer}. 
SMPLitex qualitatively outperforms the method of Lazova \textit{et al.}~\cite{lazova2019360}, which is based on a GAN inpainting network that is unable to output high-quality garment details that are visible in the input image. 
In contrast, SMPLitex textures exhibit fine details such as wrinkles and facial attributes, while also extrapolating well to the occluded parts of the body.

\vspace{-10pt}
\paragraph*{Evaluation on Market-1501~\cite{zheng2015scalable}.}
This dataset consists of a large collection of low-resolution images ($64\times128$ pixels) of 1501 different subjects, captured in an urban scenario.
It was originally proposed to evaluate person re-identification tasks, but Xu \textit{et al.}~\cite{xu2021texformer} extended their use to evaluate human texture estimation.
The idea is to estimate the texture of a subject on an image and test the fidelity of the recovered texture by rendering and comparing it to another image of the same subject captured from another viewpoint.
We use the same test set and evaluation protocol defined by Xu \textit{et al.}~\cite{xu2021texformer}.

\begin{wraptable}[13]{r}{0.48\textwidth}
    \centering
\vspace{-5pt}
\begin{tabular}{lcc}
\toprule
                & SSIM $\uparrow$    & LPIPS $\downarrow$       \\ \midrule
CMR \cite{kanazawa2018learning}       & 0.7142        & 0.1275                              \\ \hline\addlinespace[3pt]
HPBTT \cite{zhao2020human}       & 0.7420        & 0.1168                              \\ \hline\addlinespace[3pt]
RSTG \cite{wang2019reidsupervised}       & 0.6735        & 0.1778                              \\ \hline\addlinespace[3pt]
TexGlo \cite{xu20213tpami}       & 0.6658        & 0.1776                              \\ \hline\addlinespace[3pt]
TexFormer \cite{xu2021texformer}       & 0.7422        & 0.1154                              \\ \hline\addlinespace[3pt]
SMPLitex (ours) & \textbf{0.8648} & \textbf{0.0695}                                 \\ 
\bottomrule\addlinespace[3pt]
\end{tabular}
\caption{Quantitative evaluation on Market-1501 \cite{zheng2015scalable}, following the protocol and evaluation code defined by Xu \textit{et al.} ~\cite{xu2021texformer}.}
\label{tab:qunatitative-market1501}
\end{wraptable}

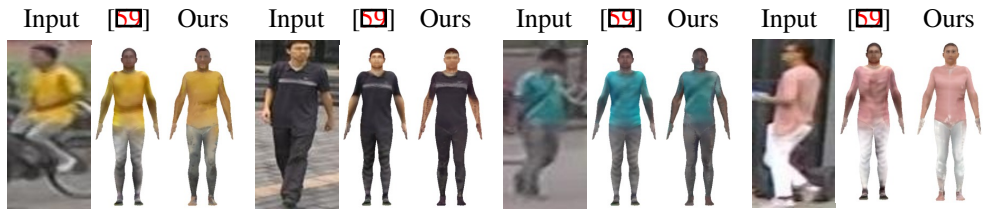
\begin{figure*}
\begin{tikzpicture}
    \node (img) at (0,0) {\includegraphics[width=\textwidth]{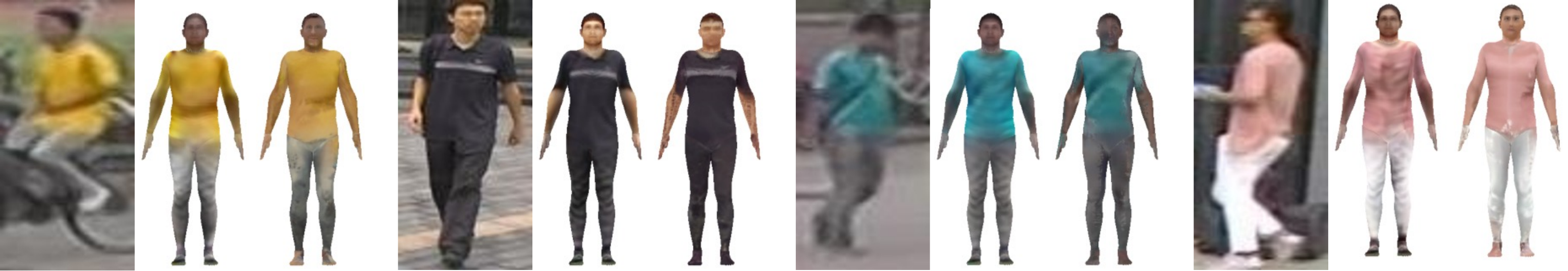}};
    \node (input1) [above, align=left, xshift = -167pt, yshift = -5pt, text height = 0.6 cm, text depth = 0.1 cm] at (img.north){Input};
    \node (xu1) [above, align=left, xshift = -138pt, yshift = -5pt, text height = 0.6 cm, text depth = 0.1 cm] at (img.north){\cite{xu2021texformer}};
    \node (ours1) [above, align=left, xshift = -110pt, yshift = -5pt, text height = 0.6 cm, text depth = 0.1 cm] at (img.north){Ours};
    \node (input2) [above, align=left, xshift = -75pt, yshift = -5pt, text height = 0.6 cm, text depth = 0.1 cm] at (img.north){Input};
    \node (xu2) [above, align=left, xshift = -46pt, yshift = -5pt, text height = 0.6 cm, text depth = 0.1 cm] at (img.north){\cite{xu2021texformer}};
    \node (ours2) [above, align=left, xshift = -19pt, yshift = -5pt, text height = 0.6 cm, text depth = 0.1 cm] at (img.north){Ours};
    \node (input3) [above, align=left, xshift = 18pt, yshift = -5pt, text height = 0.6 cm, text depth = 0.1 cm] at (img.north){Input};
    \node (xu3) [above, align=left, xshift = 47pt, yshift = -5pt, text height = 0.6 cm, text depth = 0.1 cm] at (img.north){\cite{xu2021texformer}};
    \node (ours3) [above, align=left, xshift = 74pt, yshift = -5pt, text height = 0.6 cm, text depth = 0.1 cm] at (img.north){Ours}; 
    \node (input4) [above, align=left, xshift = 110pt, yshift = -5pt, text height = 0.6 cm, text depth = 0.1 cm] at (img.north){Input};
    \node (xu4) [above, align=left, xshift = 140pt, yshift = -5pt, text height = 0.6 cm, text depth = 0.1 cm] at (img.north){\cite{xu2021texformer}};
    \node (ours4) [above, align=left, xshift = 170pt, yshift = -5pt, text height = 0.6 cm, text depth = 0.1 cm] at (img.north){Ours};      
\end{tikzpicture}
\vspace{-0.6cm}
\caption{Qualitative comparison on Market-1501 dataset \cite{zheng2015scalable}. Despite not being trained for this challenging dataset of $64 \times 128$ pixel images, SMPLitex is able to infer convincing textures that are at least on par with the dataset-specific method of Xu \textit{et al.}~\cite{xu2021texformer}.}
\label{fig:comparison-with-market}
\end{figure*}

In Table \ref{tab:qunatitative-market1501} we show that our approach SMPLitex quantitatively outperforms the state-of-the-art method by Xu \textit{et al.}~\cite{xu2021texformer}.
Furthermore, in Figure \ref{fig:comparison-with-market} we present a qualitative comparison that demonstrates that SMPLitex can infer faithful textures despite the extremely low-resolution image input.
Importantly, notice that Xu \textit{et al.} is a method trained specifically on Market-1501 (\textit{i.e.}, it does not generalize well to other datasets, as we show above in the evaluation with DeepFashion). 
SMPLitex does not suffer from such an image-specific domain, while still being competitive in the Market-1501 dataset.

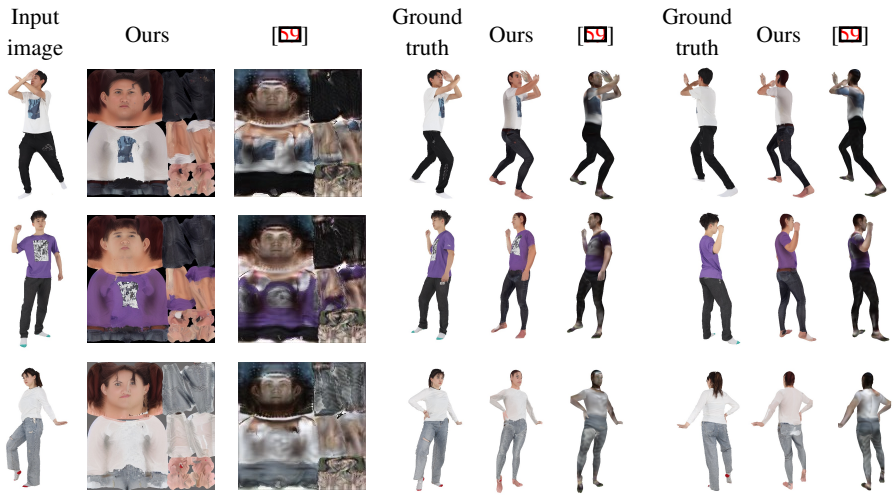
\begin{figure*}[h!]
\centering
\resizebox{0.98\columnwidth}{!}{%
\begin{tikzpicture}
    \node (img) at (0,0) {\includegraphics[trim=0 1cm 0 4cm, clip, width=\textwidth]{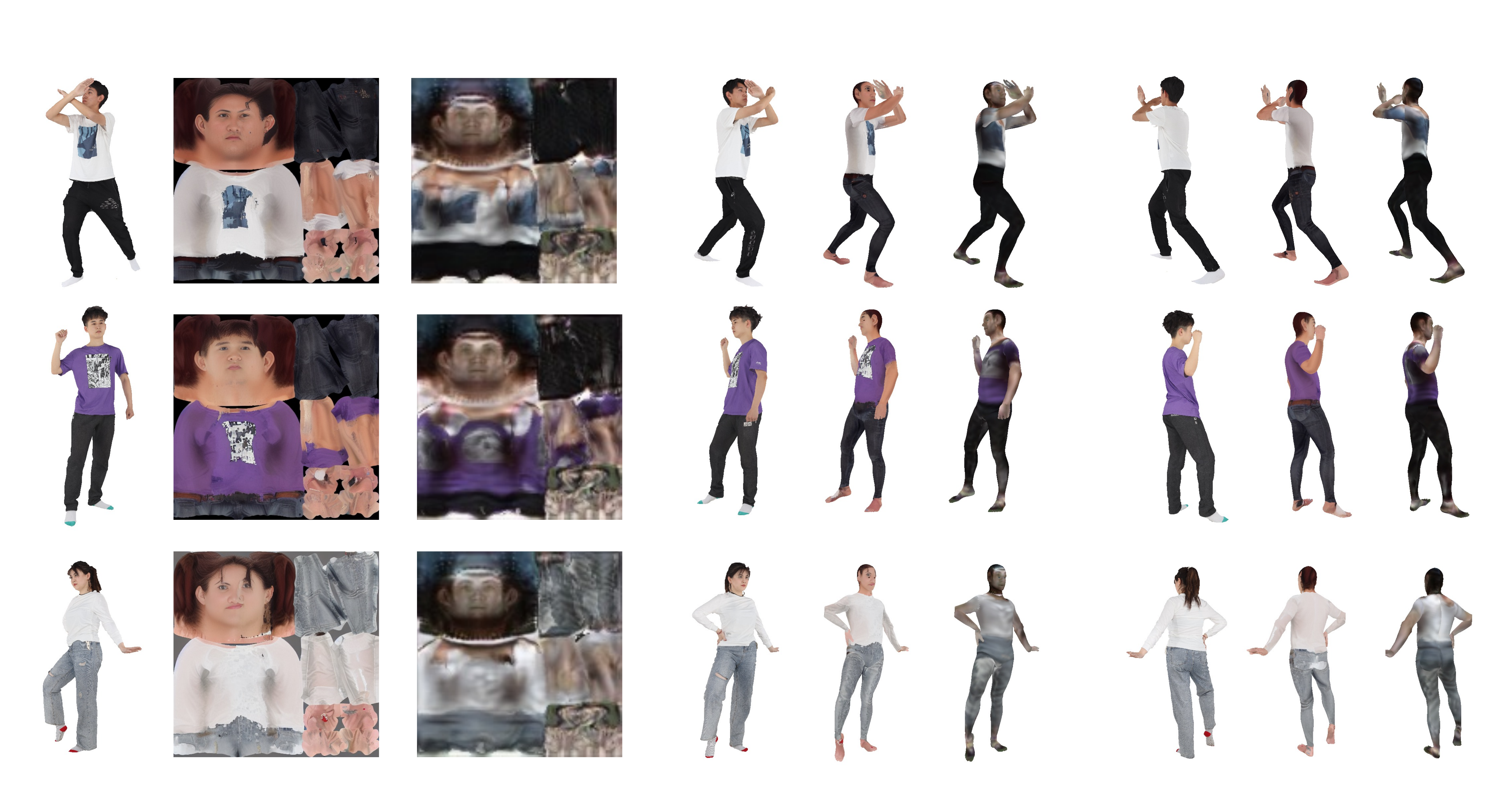}};
    \node (input1) [above, align=center, xshift = -162pt, yshift = -5pt, text height = 0.6 cm, text depth = 0.1 cm] at (img.north){\small{Input}\\\small{image}};
    \node (xu1) [above, align=left, xshift = -118pt, yshift = 0pt, text height = 0.6 cm, text depth = 0.1 cm] at (img.north){\small{Ours}};
    \node (ours1) [above, align=left, xshift = -62pt, yshift = 0 pt, text height = 0.6 cm, text depth = 0.1 cm] at (img.north){\small{\cite{xu2021texformer}}};
    \node (input2) [above, align=center, xshift = -8pt, yshift = -5pt, text height = 0.6 cm, text depth = 0.1 cm] at (img.north){\small{Ground}\\\small{truth}};
    \node (xu2) [above, align=left, xshift = 25pt, yshift = 0 pt, text height = 0.6 cm, text depth = 0.1 cm] at (img.north){\small{Ours}};
    \node (ours2) [above, align=left, xshift = 58pt, yshift = 0pt, text height = 0.6 cm, text depth = 0.1 cm] at (img.north){\small{\cite{xu2021texformer}}};
    \node (input3) [above, align=center, xshift = 98pt, yshift = -5pt, text height = 0.6 cm, text depth = 0.1 cm] at (img.north){\small{Ground}\\\small{truth}};
    \node (xu3) [above, align=left, xshift = 130pt, yshift = 0pt, text height = 0.6 cm, text depth = 0.1 cm] at (img.north){\small{Ours}};
    \node (ours3) [above, align=left, xshift = 158pt, yshift = 0pt, text height = 0.6 cm, text depth = 0.1 cm] at (img.north){\small{\cite{xu2021texformer}}}; 
\end{tikzpicture}
}
\vspace{-0.4cm}
\caption{Multi-view consistency evaluation on THUman2.0 dataset. Using the input image on the left, we show output textures by SMPLitex (ours) and \cite{xu2021texformer}. Next, we show 2 validation ground truth viewpoints and demonstrate that our renders closely match the ground truth.}
\label{fig:comparison-with-THUman}
\end{figure*}

\begin{wraptable}[8]{r}{0.48\textwidth}
    \centering
    \vspace{5pt}
\begin{tabular}{lcc}
\toprule
& SSIM $\uparrow$    & LPIPS $\downarrow$       \\ \midrule
TexFormer \cite{xu2021texformer}       & 0.8761        & 0.1223                              \\ \hline\addlinespace[3pt]
SMPLitex (ours) & \textbf{0.8829} & \textbf{0.1067}                               \\ 
\bottomrule\addlinespace[3pt]
\end{tabular}
\caption{Quantitative evaluation on THUman2.0 \cite{tao2021function4d}.}
\label{tab:qunatitative-thuman2}
\end{wraptable}
\paragraph*{Evaluation on THUman2.0 \cite{tao2021function4d}.}
To further evaluate our approach, we contribute with a new evaluation protocol for 3D texture inference from a single image.
To this end, we leverage the THUmans2.0~\cite{tao2021function4d} dataset, a high-quality collection of 3D scans with SMPL pose labels, and generate a test set consisting of a render of each scan from a specific viewpoint.
We then compare the estimated texture rendered from a different viewpoint with ground truth renders of the scan and compute the pixel similarity.  
Figure \ref{fig:comparison-with-THUman} presents qualitative results of this evaluation, demonstrating that our estimated textures closely match the ground truth images from different viewpoints.
Since we have 3D ground truth scans, we are able to compute camera-space pixel-based errors of the rendered textures. 
Table \ref{tab:qunatitative-thuman2} shows that SMPLitex outperforms the state-of-the-art method of Xu \textit{et al} \cite{xu2021texformer} in both SSIM and LPIPS metrics in this multi-view evaluation protocol on high-resolution images.

\vspace{20pt}
\section{Conclusions and Limitations}
We have presented SMPLitex, a generative model for 3D human appearance that enables the estimation of 3D human textures from single images.
SMPLitex leverages recent image diffusion models for 2D image synthesis and uses pixel-to-surface correspondence estimation to bridge the gap between 2D images and 3D surfaces.
By conditioning the diffusion model to the visible pixels of a human in a single view, SMPLitex is able to synthesize a complete texture map of the subject, outperforming current methods based on GANs or VAEs.

Despite the convincing quality of the results, SMPLitex suffers from limitations as well. If the subject on the input image is significantly occluded or if surface-to-pixel correspondence fails, SMPLitex sampling is weakly conditioned hence it can potentially generate textures that do not match well the subject.
Similarly, when sampling the model with text, if prompts are too general or not related to humans, output textures can exhibit unrealistic facial or body features such as deformed faces or missing limbs.

\paragraph{Acknowledgments.}
This work has been partially funded by: the European Union’s Horizon 2020 research and innovation program under grant agreement No 899739 (H2020-FETOPEN-2018-2020 CrowdDNA project); and by the Universidad Rey Juan Carlos through the Distinguished Researcher position INVESDIST-04 under the call from 17/12/2020.
\bibliography{egbib}
\end{document}